\def\year{2021}\relax
\documentclass[letterpaper]{article} 
\usepackage{aaai21}  
\usepackage{times}  
\usepackage{helvet} 
\usepackage{courier}  
\usepackage[hyphens]{url}  
\usepackage{graphicx} 
\usepackage{natbib}  
\usepackage{caption} 

\usepackage{todonotes}
\usepackage{xcolor}

\usepackage{tabularx}
\usepackage{booktabs}
\usepackage{comment,soulutf8}
\sethlcolor{green}
\usepackage{amssymb}
\usepackage{paralist}
\usepackage{tikz}
\usepackage{subfig}
\usepackage[dash,dot]{dashundergaps}
\usepackage{amsmath}

\usepackage{footmisc}

\newcommand{\hlg}[2][green]{{\sethlcolor{#1}\hl{#2}}}
\newcommand{\hlr}[2][orange]{{\sethlcolor{#1}\hl{#2}}}

\usepackage{pifont}
\newcommand{\cmark}{\ding{51}}%
\newcommand{\xmark}{\ding{55}}%

\nocopyright
\title{{\bf{HateXplain}}: A Benchmark Dataset for Explainable Hate Speech Detection\thanks{Accepted at AAAI 2021.}}

\author {

        Binny Mathew\textsuperscript{\rm 1}\thanks{Equal Contribution}, 
        Punyajoy Saha\textsuperscript{\rm 1}\footnotemark[2], 
        Seid Muhie Yimam\textsuperscript{\rm 2} \\
        Chris Biemann\textsuperscript{\rm 2},  
        Pawan Goyal\textsuperscript{\rm 1},  
        Animesh Mukherjee\textsuperscript{\rm 1} \\
}
\affiliations {
    \textsuperscript{\rm 1} Indian Institute of Technology, Kharagpur, India \\
    \textsuperscript{\rm 2} Universität Hamburg, Germany \\
    binnymathew@iitkgp.ac.in, punyajoys@iitkgp.ac.in, yimam@informatik.uni-hamburg.de\\
    biemann@informatik.uni-hamburg.de, pawang@cse.iitkgp.ac.in, animeshm@cse.iitkgp.ac.in
    
}

\begin{document}

\maketitle

\begin{abstract}
Hate speech is a challenging issue plaguing the online social media.
While better models for hate speech detection are continuously being developed, there is little research on the \textit{bias} and \textit{interpretability} aspects of hate speech. In this paper, we introduce {HateXplain}, the first benchmark hate speech dataset covering multiple aspects of the issue. Each post in our dataset is annotated from three different perspectives: the basic, commonly used 3-class
\textit{classification} (i.e., hate, offensive or normal), the \textit{target community} (i.e., the community that has been the victim of hate speech/offensive speech in the post), and the \textit{rationales}, i.e., the portions of the post on which their labelling decision (as hate, offensive or normal) is based. 
We utilize existing state-of-the-art models and observe that even models that perform very well in classification do not score high on explainability metrics like model \textit{plausibility} and \textit{faithfulness}. We also observe that models, which utilize the human rationales for training, perform better in reducing unintended bias towards target communities.
We have made our code and dataset public\footnote{\label{dataset_link}\url{https://github.com/punyajoy/HateXplain}} for other researchers.

\textbf{Disclaimer}: The article contains material that many will find offensive or hateful; however this cannot be avoided owing to the nature of the work.
\end{abstract}

\section{Introduction}
The increase in online hate speech is a major cultural threat, as it already resulted in crime against minorities, see e.g.~\cite{williams2019hate}. To tackle this issue, there has been a rising interest in  hate speech detection to expose and regulate this phenomenon. Several hate speech datasets \cite{ousidhoum2019multilingual,qian2019learning,de2018hate,sanguinetti2018italian}, models \cite{zhang2018detecting,mishra2018author,qian2018leveraging,qian2018hierarchical}, and shared tasks \cite{basile2019semeval,bosco2018overview} have been made available in the recent years by the community, towards the development of automatic hate speech detection.

\begin{table*}[t]
\centering
\scriptsize
\begin{tabular}{lll}
\toprule
Model & Text & Label \\
\midrule
Human Annotator & The \hlg{jews} are again using \hlg{holohoax} as an excuse to \hlg{spread} \hlg{their} \hlg{agenda} . \hlg{Hilter} should have \hlg{eradicated} them & HS \\ \midrule
CNN-GRU & The \hlg{jews} are again using \hlg{holohoax} \hlr{as} an excuse to spread \hlg{their} \hlg{agenda} . Hilter \hlr{should} \hlr{have} eradicated them & HS \\
BiRNN & \hlr{The} \hlg{jews} \hlr{are} again \hlr{using} holohoax as an excuse to spread their \hlg{agenda} . \hlg{Hilter} \hlr{should} have eradicated them & HS \\
BiRNN-Attn & The \hlg{jews} are again \hlr{using} holohoax as \hlr{an} \hlr{excuse} \hlr{to} spread \hlg{their} agenda . \hl{Hilter} should have eradicated them & HS \\
BiRNN-HateXplain & The jews are \hlr{again} \hlr{using} holohoax \hlr{as} an excuse to spread \hlg{their} agenda . \hlg{Hilter} \hlr{should} have \hlg{eradicated} them & HS \\
BERT & \hlr{The} \hlg{jews} are again using \hlg{holohoax} as an excuse to \hlg{spread} their \hlg{agenda} . Hilter should \hlr{have} \hlg{eradicated} them & OF \\
BERT-HateXplain & \hlr{The} \hlg{jews} \hlr{are} again using \hlg{holohoax} as an \hlr{excuse} to spread their \hlg{agenda} . Hilter should \hlr{have} eradicated them & OF \\
\bottomrule
\end{tabular}
\caption{Example of the rationales predicted by different models compared to human annotators. The \hlg{green highlight} marks tokens that the human annotator and model  found important for the prediction. The \hlr{orange highlight} marks tokens which the model found important, but the human annotators did not.}
\label{tab:example_model_rational}
\end{table*}

While many models have claimed to achieve state-of-the-art performance on some datasets, they fail to generalize \cite{arango2019hate,grondahl2018all}. The models may classify comments that refer to certain commonly-attacked identities (e.g., gay, black, muslim)  as  toxic  without  the  comment  having  any  intention of being toxic~\cite{dixon2018measuring,borkan2019nuanced}.
 A large prior on certain trigger vocabulary leads to biased predictions that may discriminate against particular groups who are already the target of such abuse~\cite{sap2019risk,davidson2019racial}. Another issue with the current methods is the lack of explanation about the decisions made. With hate speech detection models becoming increasingly complex, it is getting difficult to explain their decisions~\cite{bengio2017deep}.
Laws such as General Data Protection Regulation (GDPR~\cite{EU_GDPR_2016}) in Europe have recently established a ``right to explanation''. This calls for a shift in perspective from performance based models to interpretable models. 
In our work, we approach model explainability by learning the target classification and the reasons for the human decision jointly, and also to their mutual improvement.

We therefore have compiled a dataset that covers multiple aspects of hate speech. We collect posts from Twitter\footnote{\url{https://twitter.com/}} and Gab\footnote{\url{https://gab.com/}}, and ask Amazon Mechanical Turk (MTurk) workers to annotate these posts to cover three facets. In addition to classifying each post into hate, offensive, or normal speech, annotators are asked to select the target communities mentioned in the post. Subsequently, the annotators are asked to highlight parts of the text that could justify their classification decision\footnote{In case the post is classified as normal, the annotators does not need to highlight any span.}. The notion of justification, here modeled as `human attention', is very broad with many possible realizations~\cite{lipton2016mythos,doshi2017towards}. In this paper, we specifically focus on using \textit{rationales}, i.e., snippets of text from a source text that support a particular categorization. Such rationales have been used in commonsense explanations~\cite{rajani-etal-2019-explain}, e-SNLI~\cite{camburu2018snli} and several other tasks~\cite{deyoung2019eraser}. If these rationales are good reasons for decisions, then models guided towards these in training could be made more human-decision-taking-like. 

Consider the examples in Table~\ref{tab:example_model_rational}. The first row shows the tokens (`rationales') that were selected by human annotators which they believe are important for the classification. The next six rows show the important tokens (using LIME~\cite{ribeiro2016should}), which helped
various models in the classification. We observe that even when the model is making the correct prediction (hate speech -- HS in this case), the reason (`rationales') for this varies across models. In case of BERT, we observe that it attends to several of the tokens that human annotators deemed important, but assigns the wrong label (offensive speech - OF).

In summary, we introduce {\bf{HateXplain}}, the first benchmark dataset for hate speech with word and phrase level span annotations that capture human rationales for the labeling. Using MTurk, we collect a large dataset of around 20K posts and annotate them to cover three aspects of each post. We use several models on this dataset and observe that while they show a good model performance, they do not fare well in terms of model interpretability/explainability. We also observe that providing these rationales as input during training helps in improving a model's performance and reducing the unintended bias. We believe that this dataset would serve as a fundamental source for the future hate speech research.

\section{Related work}

\subsection{Hate speech}
The public expression of hate speech affects the devaluation of minority members \cite{greenberg1985effect} and such frequent and repetitive exposure to hate speech could increase an individual's outgroup prejudice \cite{soral2018exposure}. Real world violent events could also lead to increased hate speech in online space~\cite{DBLP:conf/icwsm/Olteanu0BV18}. To tackle this, various methods have been proposed for hate speech detection~\cite{burnap2016us,ribeiro2018characterizing,zhang2018detecting,qian2018hierarchical}.
The recent interest in hate speech research has led to the release of datasets in multiple languages \cite{ousidhoum2019multilingual,sanguinetti2018italian} along with different computational approaches to combat online hate~\cite{qian2019benchmark,mathew2019thou,aluru2020deep}.

A recurrent issue with the majority of previous research is that many of them tend to conflate hate speech and abusive/offensive\footnote{We have used the terms offensive and abusive interchangeably in our paper as they are arguably very similar~\cite{founta2018large}.} language \cite{davidson2017automated}. 
Some of the works have combined offensive and hate language under a single concept, while very few works, such as \cite{davidson2017automated,founta2018large} and \citet{van2019hate} have attempted to separate  offensive from hate speech. We argue that this, although subjective, is an important aspect as there are lots of messages that are offensive but do not qualify as hate speech. For example, consider the word \textit{`nigga'}. The word is used everyday in online language by the African American community~\cite{del2017hate}. Similarly, words like \textit{hoe} and \textit{bitch} are used commonly in rap lyrics. Such language is prevalent on social media \cite{wang2014cursing} and any hate speech detection system should include these for the system to be usable. To this end, we have assumed that a given text can belong to one of the three classes: hate, offensive, normal. We have adopted the classes based on the work of~\citet{davidson2017automated}. 
Table~\ref{tab:hatespeech_dataset_comparison} provides a comparison between some hate speech datasets.

\begin{table*}[tbh]
\centering
\scriptsize
\begin{tabular}{llll ll}
\toprule
Dataset         & Labels                         & Total Size & Language & Target Labels? & Rationales? \\
\midrule
\citet{waseem2016hateful} & racist, sexist, normal & 16,914        & English  & \xmark                      & \xmark         \\

\citet{davidson2017automated} & Hate Speech, Offensive, Normal & 24,802        & English  & \xmark                      & \xmark         \\
\citet{founta2018large} & Abusive, Hateful, Normal, Spam & 80,000        & English  & \xmark                      & \xmark         \\

\citet{ousidhoum2019multilingual} & Labels for five different aspects & 13,000        & English, French, Arabic  & \cmark                      & \xmark         \\
{\bf{HateXplain}} (Ours) & Hate Speech, Offensive, Normal & 20,148        & English  & \cmark                      & \cmark\\
                \bottomrule
\end{tabular}
\caption{Comparison of different hate speech datasets.}
\label{tab:hatespeech_dataset_comparison}
\end{table*}

\subsection{Explainability/Interpretability}

\citet{zaidan2007using} introduced the concept of using \textit{rationales}, in which human annotators would highlight a span of text that could support their labeling decision. The authors utilized these enriched rationale annotation on a smaller set of training data, which helped to improve sentiment classification. \citet{yessenalina2010automatically} built on this work and developed methods that automatically generate rationales. \citet{lei2016rationalizing} also proposed an encoder-generator framework, which provides quality rationales without any annotations.

In our paper, we utilize the concept of \textit{rationales} and provide the first benchmark hate speech dataset with human level explanations. 
We have made our model and dataset public\footref{dataset_link} for other researchers.

\section{Dataset collection and annotation strategies}
In this section, we provide the annotation strategies we have followed, the dataset selection approaches used, and the statistics of the collected dataset.

\subsection{Dataset sampling}

We collect our dataset from sources where previous studies on hate speech have been conducted: \textbf{Twitter}~\cite{davidson2017automated,fortuna2018survey} and \textbf{Gab}~\cite{lima2018inside,mathew2020hatebegets,zannettou2018gab}. Following the existing literature, we build a corpus of posts (tweets and gab posts) using lexicons. We combined the lexicon set provided by \citet{davidson2017automated}, \citet{ousidhoum2019multilingual}, and \citet{mathew2019spread} to generate a single lexicon. For Twitter, we filter the tweets from the 1\% randomly collected tweets in the time period Jan-2019 to Jun-2020. In case of Gab, we use the dataset provided by \citet{mathew2019spread}.  We do not consider reposts and remove duplicates. We also ensure that the posts do not contain links, pictures, or videos as they indicate additional information that might not be available to the annotators. However, we do not exclude the emojis from the text as they might carry important information for the hate and offensive speech labeling task. The posts were anonymized by replacing the usernames with \textless user\textgreater token.

\subsection{Annotation procedure}
We use Amazon Mechanical Turk (MTurk) workers for our annotation task. Each post in our dataset contains three types of annotations. First, whether the text is a hate speech, offensive speech, or normal. Second, the target communities in the text. Third, if the text is considered as hate speech, or offensive by majority of the annotators, we further ask the annotators to annotate parts of the text, which are words or phrases that could be a potential reason for the given annotation. These additional span annotations allow us to further explore how hate or offensive speech manifests itself.

\subsubsection{Target group annotation}
The primary goal of the annotation task is to determine whether a given text is hateful, offensive, or neither of the two, i.e. normal. As noted above, we also get span annotations as reasons for the label assigned to a post (hateful or offensive). To further enrich the dataset, we ask the workers to decide the groups that the hate/offensive speech is targeting. Table~\ref{tab:targets} lists the target groups we have identified~\footnote{The data uses the label ``homosexual'' as defined at collection time instead of gay; other sexual and gender orientation categories have been pruned from the data due to low incidence; the published version of the paper wrongly mentions the LGBTQ category.}.

\begin{table}[htbp]
\centering
\resizebox{1.0\linewidth}{!}{
\begin{tabular}{|l|l|}
\hline
Target groups &  Categories\\
\hline
Race & African, Arabs, Asians, Caucasian, Hispanic\\
Religion     & Buddhism, Christian, Hindu, Islam, Jewish \\
Gender & Men, Women\\
Sexual Orientation & Heterosexual, Gay\\
Miscellaneous & Indigenous, Refugee/Immigrant, None, Others\\
\hline
\end{tabular}
}
\caption{Target groups considered for the annotation.}
\label{tab:targets}
\end{table}

\subsubsection{Annotation instructions and design of the interface}
Before starting the annotation task, workers are explicitly warned that the annotation task displays some hateful or offensive content.
We prepare instructions for workers that clearly explain the goal of the annotation task, how to annotate spans and also include a definition for each category. We provided multiple examples with classification, target community and span annotations to help the annotators understand the task. To further ensure high quality dataset, we use built-in MTurk qualification requirements, namely the \textit{HIT Approval Rate} (95\%) \textit{for all Requesters' HITs} and the \textit{Number of HITs Approved} (5,000) requirements.

\subsection{Dataset creation steps}
For the dataset creation, we first conducted a pilot annotation study followed by the main annotation task.

\noindent\textbf{Pilot annotation}: In the pilot task, each annotator was provided with 20 posts and they were required to do the hate/offensive speech classification as well as identify the target community (if any). In order to have a clear understanding of the task, they were provided with multiple examples along with explanations for the labelling process. The main purpose of the pilot task was to shortlist those annotators who were able to do the classification accurately. We also collected feedback from annotators to improve the main annotation task. A total of 621 annotators took part in the pilot task. Out of these, 253 were selected for the main task.

\begin{table}[!t]
\centering
\footnotesize
\begin{tabular}{l|ll|l}
          & Twitter & Gab  & Total \\ \hline
Hateful   & 708    & 5,227 & 5,935  \\
Offensive & 2,328    & 3,152 & 5,480  \\
Normal    & 5,770    & 2,044 & 7,814  \\ 
Undecided & 249     & 670  & 919 \\ \hline
Total     & 9,055    & 11,093 & 20,148 \\

\end{tabular}
\caption{Dataset details. ``Undecided'' refers to the cases where all the three annotators chose a different class.}
\label{tab:dataset_details}
\end{table}

\noindent\textbf{Main annotation}: After the pilot annotation, once we had ascertained the quality of the annotators, we started with the main annotation task. In each round, we would select a batch of around 200 posts. Each post was annotated by three annotators, then majority voting was applied to decide the final label. The final dataset is composed of 9,055 posts from Twitter and 11,093 posts from Gab. Table \ref{tab:dataset_details} provides further details about the dataset collected. Table~\ref{tab:dataset_example} shows samples of our dataset. The Krippendorff's $\alpha$ for the inter-annotator agreement is 0.46 which is much higher than other hate speech datasets~\cite{del2017hate,ousidhoum2019multilingual}.

\noindent\textbf{Class labels}: The class label (hateful, offensive, normal) of a post was decided based on majority voting. We found 919 cases where all the three annotators chose a different class. We did not consider these posts for our analysis. 

To decide the target community of a post, we rely on majority voting. We consider that a target community is present in the post, if at least two out of the three annotators have selected the target from Table \ref{tab:targets}. We also add a filter that the community should be present in at least 100 posts. Based on this criteria, our dataset had the following ten communities: \textit{African, Islam, Jewish, Gay, Women, Refugee, Arab, Caucasian, Hispanic, Asian}. The target community information would allow researchers to delve into issues related to bias in hate speech~\cite{davidson2019racial}. In our dataset, the top three communities that are targets of hate speech are the \textit{African}, \textit{Islam}, and \textit{Jewish} community. In case of offensive speech, the top three targets are \textit{Women}, \textit{Africans}, and \textit{Gay}. These observations are in agreement with previous research~\cite{silva2016analyzing}.

For the rationales' annotation, each post that is labelled as hateful or offensive was further provided to the annotators\footnote{We tried to get the original annotator to highlight, however this was not always possible.} to highlight the rationales that could justify the final class. Each post had rationale explanations provided by 2-3 annotators. We observe that the average number of tokens highlighted per post is 5.48 for offensive speech, and 5.47 for hate speech. Average token per post in the whole dataset is 23.42. The top three content words in the hate speech rationales are \textit{nigger, kike, and moslems}, which are found in 30.02\% of all the hateful posts. 
The top three content words for the offensive highlights are \textit{retarded, bitch, and white}, which are found in 47.36\% of all the offensive posts. 

\begin{figure}[t]
    \centering
    \includegraphics[width=\linewidth]{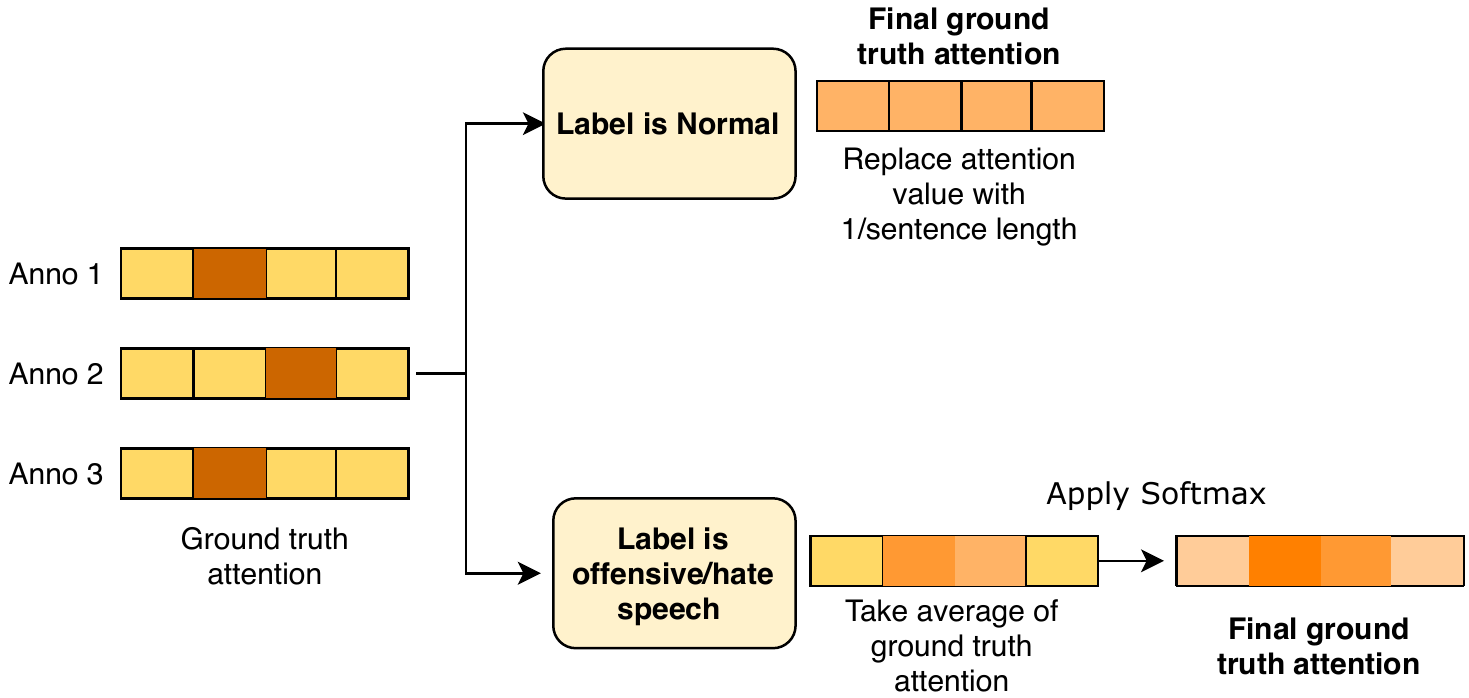}
    \caption{Ground truth attention.}
    \label{fig:ground_truth_attention}
\end{figure}

\noindent\textbf{Ground truth attention:} In order to generate the ground truth attention for the post with hate speech/offensive label, we first convert each rationale into an attention vector. This is a Boolean vector with length equal to the number of tokens in the sentence. The tokens in the rationale are indicated by a value of 1 in the attention vector. Now we take the average of the these attention vectors to represent a common ground truth attention vector for each post. The attention vectors from the attention based models usually have their sum of elements equal to 1. We normalize this common attention vector through a softmax function to generate the ground truth attention. One issue with the ground truth attention vector could be that the difference between the values of rationale and non-rationale tokens could be low. To handle this, we make use of the temperature parameter ($\tau$) in the softmax function. This allows us to make the probability distribution concentrate on the rationales. We tune this parameter using the validation set.
Finally, if the label of the post is normal, we ignore the attention vectors and replace each element in the ground truth attention with 1/(sentence length) to represent uniform distribution. We illustrate this computation in Figure \ref{fig:ground_truth_attention}.

\begin{table}[t]
\centering
\scriptsize
\begin{tabular}{ll}
\toprule
Text & Dad should have told the \hl{muzrat whore} \\
&to \hl{fuck off}, and went in anyway\\
Label & Hate \\
Targets & Islam \\ 
\midrule
Text & A \hl{nigress too dumb to fuck} has a scant \\
&chance of understanding anything beyond\\
&the size of a dick \\
Label & Hate \\
Targets & Women, African \\ 
\midrule
Text & Twitter is full of tween \hl{dikes} who think \\
&they're superior because of \hl{``muh oppression.''} \\
&News flash: No one gives a shit. \\
Label & Offensive \\
Targets & Gay \\ 

\bottomrule
\end{tabular}
\caption{Examples from our dataset. The \hl{highlighted} portion of the text represents the annotator's rationale.}

\label{tab:dataset_example}
\end{table}

\section{Metrics for evaluation}
To build the {\bf{HateXplain}} benchmark dataset, we consider multiple types of metrics to cover several aspects of hate speech. Taking inspiration from the different issues reported for hate speech classifications, we concentrate on three major types of metrics.

\subsection{Performance based metrics}
Following the standard practices, we report \textbf{accuracy}, \textbf{macro F1-score}, and \textbf{AUROC} score. These metrics would be able to evaluate the classifier performance in distinguishing among the three classes, i.e., hate speech, offensive speech, and normal.

\subsection{Bias based metrics}
The hate speech detection models could make biased predictions for particular groups who are already the target of such abuse
~\cite{sap2019risk,davidson2019racial}. For example, the sentence ``I love my niggas.'' might be classified as hateful/offensive because of the association of the word niggas with the black community. 
These unintended identity-based bias could have negative impact on the target community. To measure such unintended model bias, we rely on the AUC based metrics developed by \citet{borkan2019nuanced}. These include Subgroup AUC, Background Positive Subgroup Negative (BPSN) AUC, Background Negative Subgroup Positive (BNSP) AUC, Generalized Mean of Bias AUCs. The task here is to classify the post as \textit{toxic (hate speech, offensive)} or \textit{not (normal)}. Here, the models will be evaluated on the grounds of how much they are able to reduce the unintended bias towards a target community~\cite{borkan2019nuanced}. We restrict the evaluation to the test set only. By having this restriction, we are able to evaluate models in terms of bias reduction. Below, we briefly describe each of the metrics.

\noindent\textbf{Subgroup AUC}: Here, we select  toxic and normal posts from the test set that mention the community under consideration. The ROC-AUC score of this set will provide us with the Subgroup AUC for a community. This metric measures the model's ability to separate the toxic and normal comments in the context of the community (e.g., Asians, Gay etc.). A higher value means that the model is doing a good job at distinguishing the toxic and normal posts specific to the community.

\noindent\textbf{BPSN (Background Positive, Subgroup Negative) AUC}: Here, we 
select normal posts that mention the community and  toxic posts that do not mention the community, from the test set. The ROC-AUC score of this set will provide us with the BPSN AUC for a community. This metric measures the false-positive rates of the model with respect to a community. A higher value means that a model is less likely to confuse between the normal post that mentions the community with a toxic post that does not.

\noindent\textbf{BNSP (Background Negative, Subgroup Positive) AUC}: Here, we select toxic posts that mention the community and normal posts that do not mention the community, from the test set. The ROC-AUC score for this set will provide us with the BNSP AUC for a community. The metric measures the false-negative rates of the model with respect to a community. A higher value means that the model is less likely to confuse between a toxic post that mentions the community with a normal post without one.

\noindent\textbf{GMB (Generalized Mean of Bias) AUC}: This metric was introduced by the Google Conversation AI Team as part of their Kaggle competition\footnote{\url{https://www.kaggle.com/c/jigsaw-unintended-bias-in-toxicity-classification/overview/evaluation}}. This metric combines the per-identify Bias AUCs into one overall measure as $M _ { p } \left( m _ { s } \right) = \left( \frac { 1 } { N } \sum _ { s = 1 } ^ { N } m _ { s } ^ { p } \right) ^ { \frac { 1 } { p } }
$ where, $M_p$  = the $p^\textrm{th}$ power-mean function, $m_s$ = the bias metric $m$ calculated for subgroup $s$ and
$N$ = number of identity subgroups (10). We use $p=-5$ as was also done in the competition.

\noindent We report the following three metrics for our dataset.
\begin{compactitem}
    \item[-] \textbf{GMB-Subgroup-AUC}: GMB AUC with Subgroup AUC as the bias metric.
    \item[-] \textbf{GMB-BPSN-AUC}: GMB AUC with BPSN AUC as the bias metric.
    \item[-] \textbf{GMB-BNSP-AUC}: GMB AUC with BNSP AUC as the bias metric.
\end{compactitem}

\subsection{Explainability based metrics}

We follow the framework in the ERASER benchmark by~\citet{deyoung2019eraser} to measure the explainability aspect of a model. We measure this using \textit{plausibility} and \textit{faithfulness}. \textit{Plausibility} refers to how convincing the interpretation is to humans,  while \textit{faithfulness} refers to how accurately it reflects the true reasoning process of the model~\cite{jacovi2020towards}.

For completeness, we explain the metrics briefly below.

\noindent\textbf{Plausibility}
To measure the plausibility, we consider metrics for both discrete and soft selection. We report the IOU F1-Score and token F1-Score metric for the discrete case, and the AUPRC score for soft token selection~\cite{deyoung2019eraser}. 

Intersection-Over-Union (IOU) permits credit assignment for partial matches.  \citet{deyoung2019eraser} defines IOU on a token level: for two spans, it is the size of the overlap of the tokens they cover divided by the size of their union. A prediction is considered as a match if the overlap with any of the ground truth rationales is more than 0.5. We use these partial matches to calculate an F1-score (IOU F1). We also measure token-level precision and recall, and use these to derive token-level F1 scores (token F1). To measure the plausibility for soft token scoring,
 we also report the Area Under the Precision-Recall curve (AUPRC) constructed by sweeping a threshold over the token scores.

\noindent\textbf{Faithfulness}
To measure the faithfulness, we report two metrics: \textit{comprehensiveness} and \textit{sufficiency}~\cite{deyoung2019eraser}.
\begin{compactitem}
    \item[-]  \textbf{Comprehensiveness}: To measure comprehensiveness, we create a contrast example $\tilde{x}_i$, for each post $x_i$,
    where $\tilde{x}_i$ is calculated by removing the predicted rationales $r_i$\footnote{We select the top 5 tokens as the rationales. The top 5 is selected as it is the average length of the annotation span in the dataset.} from $x_i$. Let $m(x_i)_j$ be the original prediction probability provided by a model $m$ for the predicted class $j$. Then we define $m(x_i \backslash r_i)_j$ as the predicted probability of $\tilde{x}_i$ ($=x_i \backslash r_i$) by the   model $m$ for the  class $j$.
    We would expect the model prediction to be lower on removing the rationales. We can measure this as follows -- ${\text{comprehensiveness}} = m(x_i)_j - m(x_i \backslash r_i)_j$.
    A high value of comprehensiveness implies that the rationales were influential in the prediction.
    
    \item[-] \textbf{Sufficiency} measures the degree to which extracted rationales are adequate for a model to make a prediction. We can measure this as follows -- ${\text{sufficiency}} = m(x_i)_j - m(r_i)_j$.
\end{compactitem}

\begin{figure}[!th]
    \centering
    \includegraphics[width=0.7\linewidth]{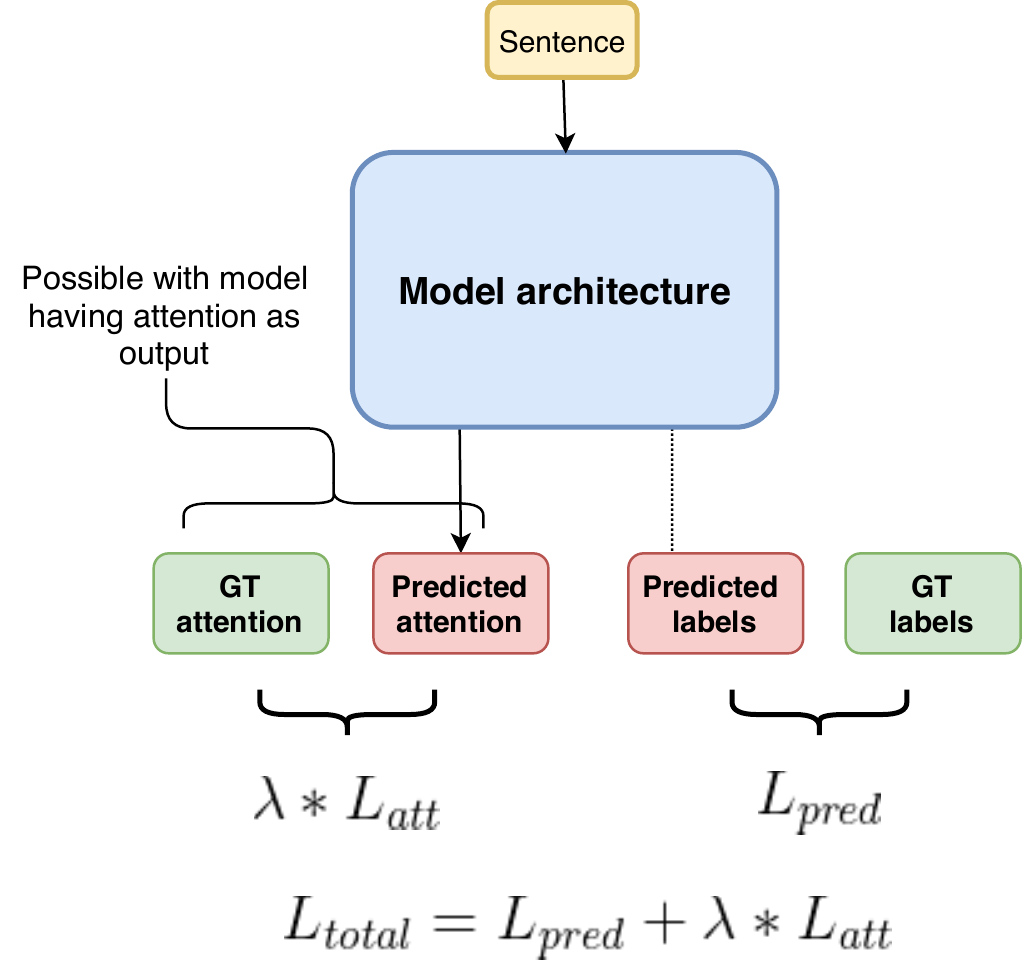}
    \caption{Representation of the general model architecture showing how the attention of the model is trained using the ground truth (GT) attention. $\lambda$ controls how much effect the attention loss has on the total loss.}
    \label{fig:model_archi}
\end{figure}

\begin{table*}[htb!]
\centering
\resizebox{1.0\linewidth}{!}{
\begin{tabular}{l| lll | lll | lll|ll}
\textbf{Model [Token Method]}                          & \multicolumn{3}{c}{\textbf{Performance}} & \multicolumn{3}{c}{\textbf{Bias}} &
\multicolumn{5}{c}{\textbf{Explainability}} \\
&&&&&&&\multicolumn{3}{c}{Plausibility}&\multicolumn{2}{c}{Faithfulness} \\
                    & Acc.$\uparrow$& Macro F1$\uparrow$& AUROC$\uparrow$& GMB-Sub.$\uparrow$& GMB-BPSN$\uparrow$& GMB-BNSP$\uparrow$& IOU F1$\uparrow$& Token F1$\uparrow$& AUPRC$\uparrow$& Comp.$\uparrow$& Suff.$\downarrow$ \\ \hline
CNN-GRU [LIME]             &0.627          &0.606              &0.793           &0.654              &0.623              &0.659              &0.167             &0.385              &0.648          &0.316          &\textbf{-0.082}  \\
BiRNN [LIME]              &0.595          &0.575              &0.767           &0.640              &0.604              &0.671              &0.162             &0.361              &0.605          &0.421          &-0.051  \\
BiRNN-Attn [Attn]  &0.621          &0.614              &0.795           &0.653              &0.662              &0.668              &0.167             &0.369              &0.643          &0.278          &0.001  \\
BiRNN-Attn [LIME]  &0.621          &0.614              &0.795           &0.653              &0.662              &0.668              &0.162             &0.386              &0.650          &0.308          &-0.075  \\
BiRNN-{\bf{HateXplain}} [Attn]&0.629       &0.629              &0.805           &0.691              &0.636              &0.674              &\textbf{0.222}    &\textbf{0.506}     &\textbf{0.841} &0.281          &0.039  \\ 
BiRNN-{\bf{HateXplain}} [LIME] &0.629      &0.629              &0.805           &0.691              &0.636              &0.674              &0.174             &0.407              &0.685          &0.343          &-0.075  \\ 
BERT [Attn]         &0.690 &0.674              &0.843           &0.762              &0.709              &0.757              &0.130             &0.497              &0.778          &0.447          &0.057\\
BERT [LIME]         &0.690 &0.674              &0.843           &0.762              &0.709              &0.757              &0.118             &0.468              &0.747          &0.436          &0.008\\
BERT-{\bf{HateXplain}} [Attn]&\textbf{0.698}        &\textbf{0.687}     &\textbf{0.851}  &\textbf{0.807}     &\textbf{0.745}     &\textbf{0.763}     &0.120             &0.411              &0.626          &0.424          &0.160   \\
BERT-{\bf{HateXplain}} [LIME]&\textbf{0.698}        &\textbf{0.687}     &\textbf{0.851}  &\textbf{0.807}     &\textbf{0.745}     &\textbf{0.763}     &0.112             &0.452              &0.722          &\textbf{0.500} &0.004   \\
\hline
\end{tabular}}%
\caption{Model performance results. To select the tokens for explainability calculation, we used attention and LIME methods.}
\label{tab:results}
\end{table*}

\section{Model details}
\label{modeldetails}
In this section, we provide details on the models used to evaluate the dataset. Each model has two versions, one where the models are trained using the ground truth class labels only (i.e., hate speech, offensive speech, and normal) and the other, where the models are trained using the ground truth attention and class labels, as shown in Figure~\ref{fig:model_archi}. For training using the ground truth attention, the model needs to output some form of vector representing attention for each token according to the model, hence, the second version is not feasible for BiRNN and CNN-GRU models\footnote{The limitation is due to the lack of an attention mechanism.}.

\subsubsection{CNN-GRU}
\citet{zhang2018detecting} used CNN-GRU to achieve state-of-the-art for multiple hate speech datasets. We modify the original architecture to include convolution 1D filters of window sizes 2, 3, 4 with each size having 100 filters. For the RNN part, we use GRU layer and finally max-pool the output representation from the hidden layers of the GRU architecture. This hidden layer is passed through a fully connected layer to finally output the prediction logits.

\subsubsection{BiRNN} For the BiRNN~\cite{schuster1997bidirectional} model, we pass the tokens in the form of embeddings to a sequential model\footnote{We experiment with LSTM and GRU.}. The last hidden state is passed through 2 fully connected layers. The output after that is used as the prediction logits. We use  dropout layers after the embedding layer and before both the fully connected layers to regularise the trained model. 

\subsubsection{BiRNN-Attention} 
This model is identical to the BiRNN model but includes an attention layer~\cite{liu2016attention} after the sequential layer. This attention layer outputs an attention vector based on a context vector which is analogous to asking \textit{``which is the most important word?''}. Weights from the attention vector are multiplied with the output hidden units from the sequential layer and added to present a final representation of the sentence. This representation is passed through two fully connected layers as in the BiRNN model. Further to train the attention layer outputs, 
we compute cross entropy loss between the attention layer output and the ground truth attention (cf. Figure~\ref{fig:ground_truth_attention} for its computation) as shown in Figure \ref{fig:model_archi}.

\subsubsection{BERT} BERT~\cite{devlin2019bert} stands for Bidirectional Encoder Representations from Transformers pre-trained on data from English language\footnote{We use the bert-base-uncased model having 12-layer, 768-hidden, 12-heads, 110M parameters.}. It is a stack of transformer encoder layers with 12 ``attention heads'', i.e., fully connected neural networks augmented with a self attention mechanism. In order to fine-tune BERT, we add a fully connected layer with the output corresponding to the \textit{CLS} token in the input. 
This \textit{CLS} token output usually holds the representation of the sentence. Next, to add \textit{attention supervision}, we try to match the attention values corresponding to the \textit{CLS} token in the final layer to the ground truth attention, so that when the final weighted representation of \textit{CLS} is generated, it would give attention to words as per the ground truth attention vector. This is calculated using a cross entropy between the attention values and the ground truth attention vector as shown in Figure \ref{fig:model_archi}.

\begin{figure*}[!tbh]%
    \centering
    \subfloat[\centering SubGroup]{{\includegraphics[width=0.3\linewidth]{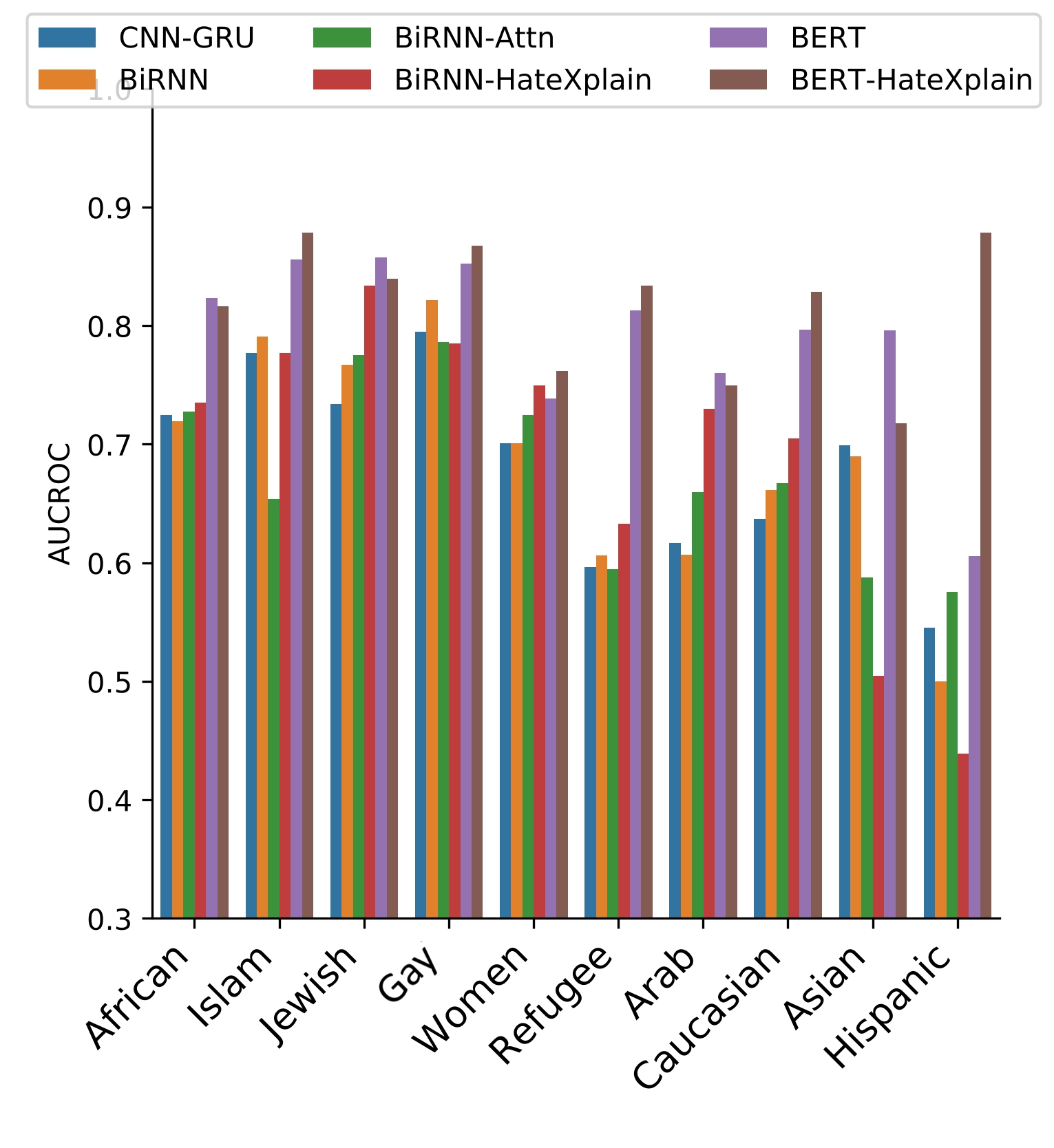} \label{fig:bias_community_wise_subgroup}}}%
    \subfloat[\centering BPSN ]{{\includegraphics[width=0.3\linewidth]{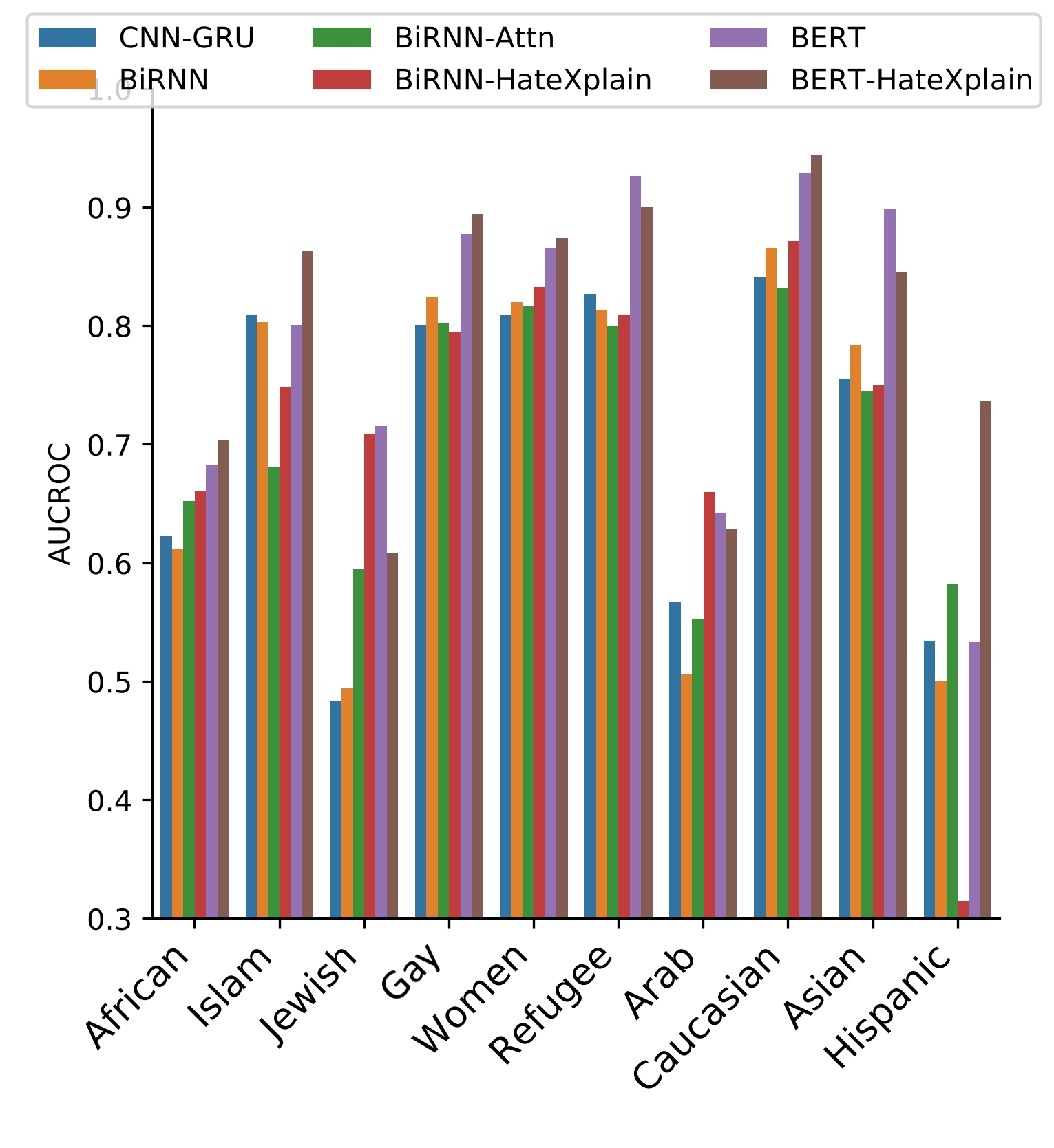}\label{fig:bias_community_wise_bpsn}}}%
    \subfloat[\centering BNSP ]{{\includegraphics[width=0.3\linewidth]{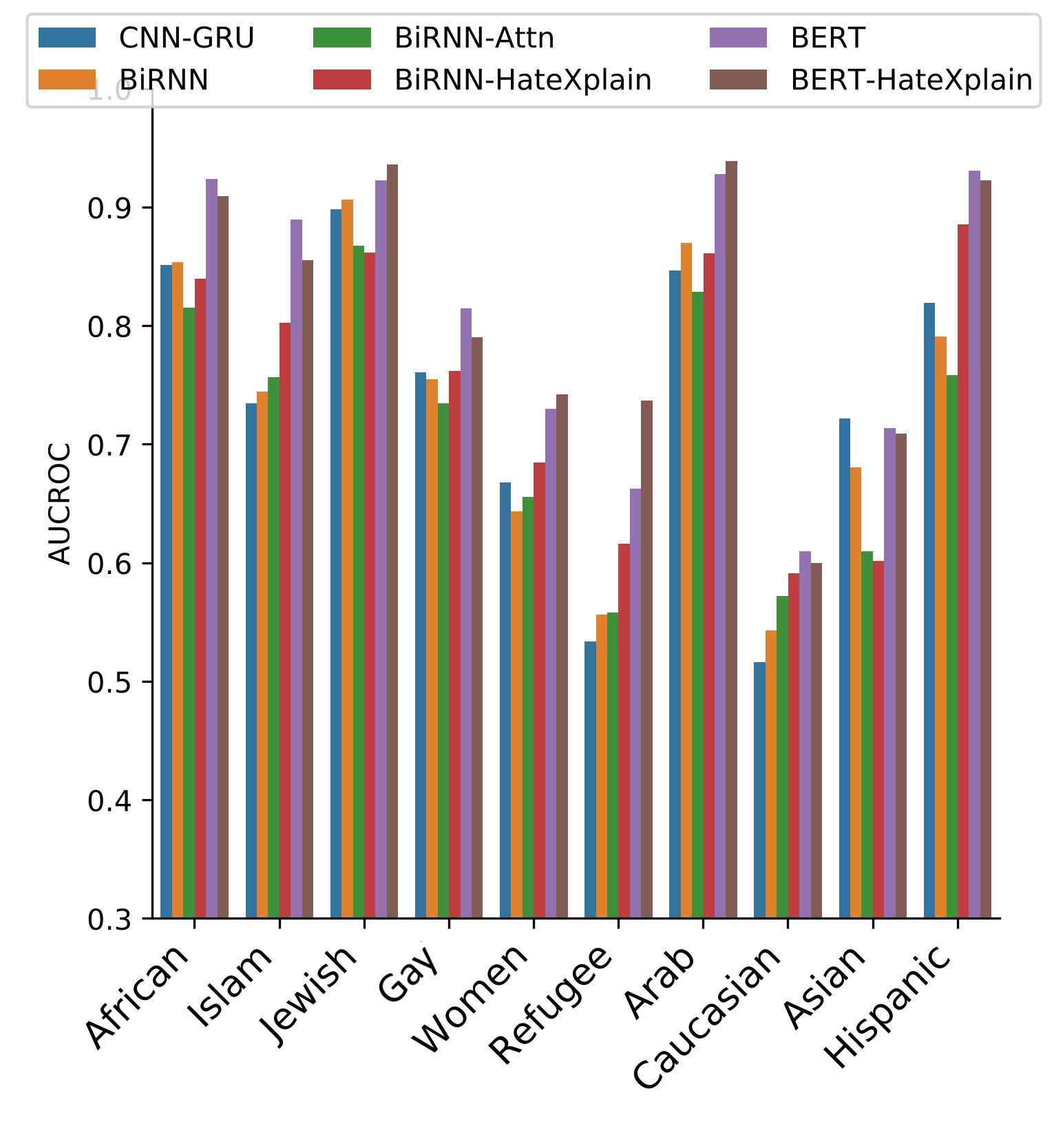} \label{fig:bias_community_wise_bnsp}}}%
    \caption{Community-wise results for each of the bias metrics.}%
    \label{fig:bias_community_wise}%
\end{figure*}

\subsection{Hyper-parameter tuning}
All the methods are compared using the same \textit{train}:\textit{development}:\textit{test} split of 8:1:1. We perform stratified split on the dataset to maintain class balance. All the results are reported on the test set and the development set is used for hyper-parameter tuning.
We use the common crawl\footnote{840B tokens, 2.2M vocab, cased, 300d vectors.} pre-trained GloVe embeddings~\cite{pennington2014glove} to initialize the word embeddings for the non-BERT models. In our models, we set the token length to 128 for faster processing of the query\footnote{Almost all the posts consist of less than 128 tokens in the data.}. We use Adam~\cite{kingma2014adam} optimizer and find the learning rate to 0.001 for the non-BERT models and 2e-5 for BERT models using the development set. The RNN models prefer LSTM as the sequential layer with hidden layer size of 64 for BiRNN with attention and 128 for BiRNN.  We use dropouts at different levels of the model. The regulariser $\lambda$ controls how much effect the attention loss has on the total loss as in Figure \ref{fig:model_archi}. Optimum performance occurs with $\lambda$ being set to 100 for BiRNN with attention and BERT with attention in the supervised setting\footnote{Please note that our selection of the best hyper-parameter was based on the model performance, which is in lines with what is suggested in the literature. One could have a variant where the model is optimized for the best explainability. This dataset gives researchers the flexibility to choose best parameters based on plausibility and/or faithfulness.}.

\section{Results}
 We report the main results obtained in Table~\ref{tab:results}.

\noindent\textbf{Performance}: We observe that models  utilizing the human rationales as part of the training (BiRNN-{\bf{HateXplain} [LIME \& Attn]}, BERT-{\bf{HateXplain} [LIME \& Attn]}\footnote{\textless model\textgreater-{\bf{HateXplain}} denotes the models where we use supervised attention using ground truth attention vector.}) are able to perform slightly better in terms of the performance metrics. 
BiRNN-{\bf{HateXplain} [LIME \& Attn]} has improved score for all plausibitliy metrics and comprehensiveness as compared to BiRNN-Attn [LIME \& Attn]. In case of BERT-{\bf{HateXplain} [LIME]}, the faithfulness scores have improved as compared to other BERT models. However, the plausibility scores have decreased.

\noindent\textbf{Bias}: Similar to performance, models that utilize the human rationales as part of the training are able to perform better in reducing the unintended model bias for all the bias metrics. We observe that presence of community terms within the rationales is effective in reducing the unintended bias.
We also looked at the model bias for each individual community in Figure~\ref{fig:bias_community_wise}. 
Figure~\ref{fig:bias_community_wise_subgroup} reports the community wise subgroup AUCROC. We observe that while the GMB-Subgroup metric reports $\sim$0.8 AUROC, the score for individual community has large variations. Target communities like Asians have scores $\sim$0.7, even for the best model. Communities like Hispanic seem to be biased toward having more false positives. Models like BERT-\textbf{HateXplain} seem to be able to handle this bias much better than other models. Future research on hate speech, should consider the impact of the model performance on individual communities to have a clear understanding on the impact.

\noindent\textbf{Explainability}: We observe that models such as BERT-{\bf{HateXplain} [LIME \& Attn]}, which attain the best scores in terms of performance metrics and bias, do not perform well in terms of plausibility explainability metrics. In fact, BERT-{\bf{HateXplain} [Attn]} has the worst score for sufficiency as compared to other models. BERT-{\bf{HateXplain} [LIME]} seems to be the best model for comprehensiveness metric. 
For plausibility metrics, we observe BiRNN-{\bf{HateXplain} [Attn]} to have the best scores. For sufficiency, CNN-GRU seems to be doing the best. For the token method, LIME seems to be generating more faithful results as compared to attention. These are in agreement with \citet{deyoung2019eraser}. Overall, we observe that a model's performance metric alone is not enough. Models with slightly lower performance, but much higher scores for plausibility and faithfulness might be preferred depending on the task at hand. The \textbf{HateXplain} dataset could be a valuable tool for researchers to analyze and develop models that provide more explainable results. 

\noindent\textbf{Variations with $\lambda$}: We measure the effect of $\lambda$ on model performance (macro F1 and AUROC) and explainability (token F1, AUPRC, comp., and suff.). We experiment with BiRNN-{\bf{HateXplain} [Attn]} and BERT-{\bf{HateXplain} [Attn]}. Increasing the value of $\lambda$ improves the model performance, plausability, and sufficienty while degrading comprehensiveness.

\section{Limitations of our work}
Our work has several limitations. First is the lack of external context. In our current models, we have not considered any external context such as profile bio, user gender, history of posts etc., which might be helpful in the classification task. Also, in this work we have focused on the English language. It does not consider multilingual hate speech into account.

\section{Conclusion and future work}
In this paper, we have introduced {\bf{HateXplain}}, a new benchmark dataset\footref{dataset_link} for hate speech detection. The dataset consists of 20K posts from Gab and Twitter. Each data point is annotated with one of the hate/offensive/normal labels, target communities mentioned, and snippets (rationales) of the text marked by the annotators who support the label. We test several state-of-the-art models on this dataset and perform evaluation on several aspects of the hate speech detection. Models that perform very well in classification cannot always provide plausible and faithful rationales for their decisions. 

As part of the future work, we plan to incorporate existing hate speech datasets~\cite{davidson2017automated,ousidhoum2019multilingual,founta2018large} to our {\bf{HateXplain}} framework.

\bibliography{main}

\section{Appendix}

\section{Interface design}

\begin{figure}[!tbh]
 \includegraphics[width=\linewidth]{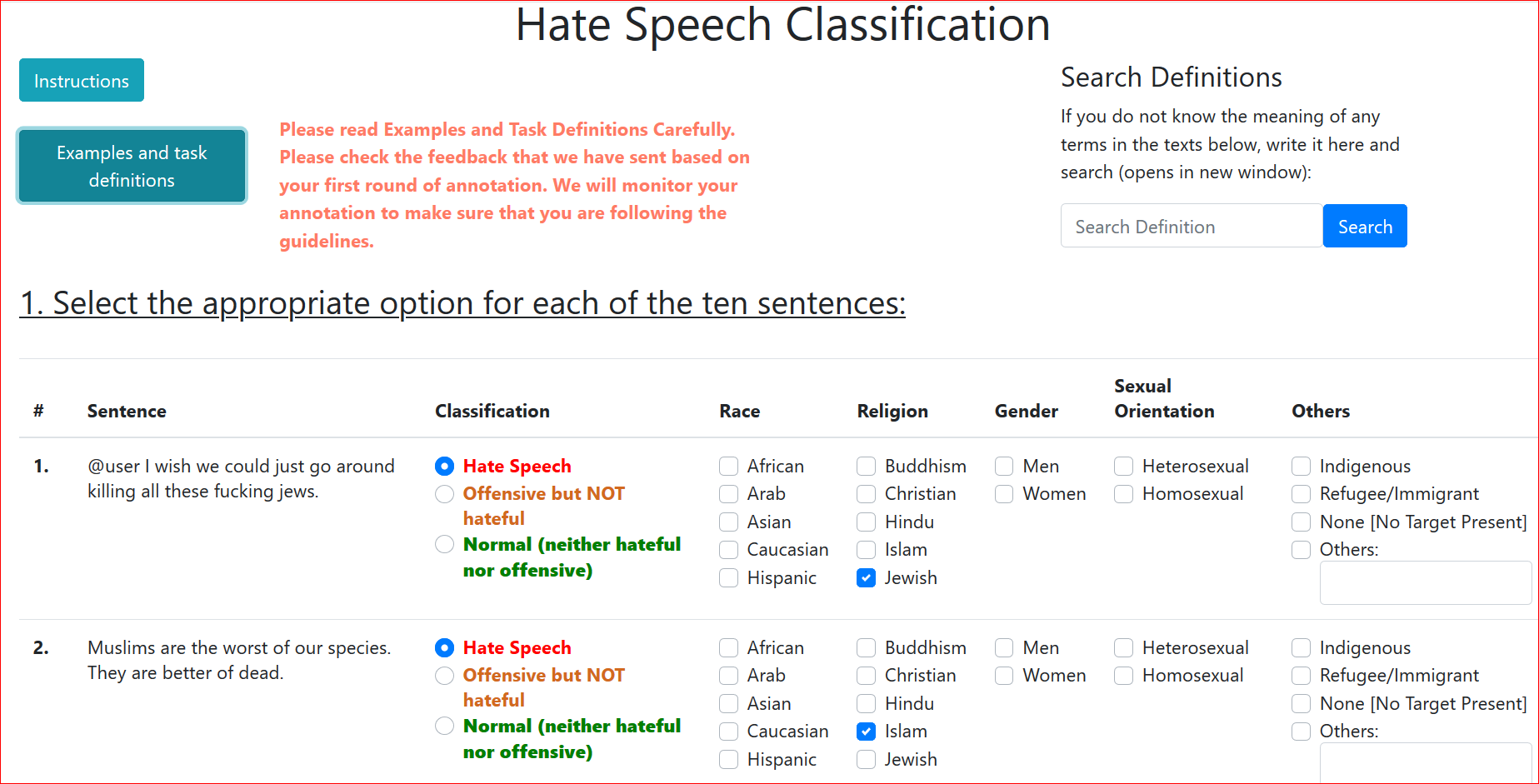}
 \caption{The classification interface. The annotator is provided with 20 text messages and asked to selected the correct type and target of the message.} 
 \label{fig:anno1}
\end{figure}

\begin{figure}[!tbh]
 \includegraphics[width=\linewidth]{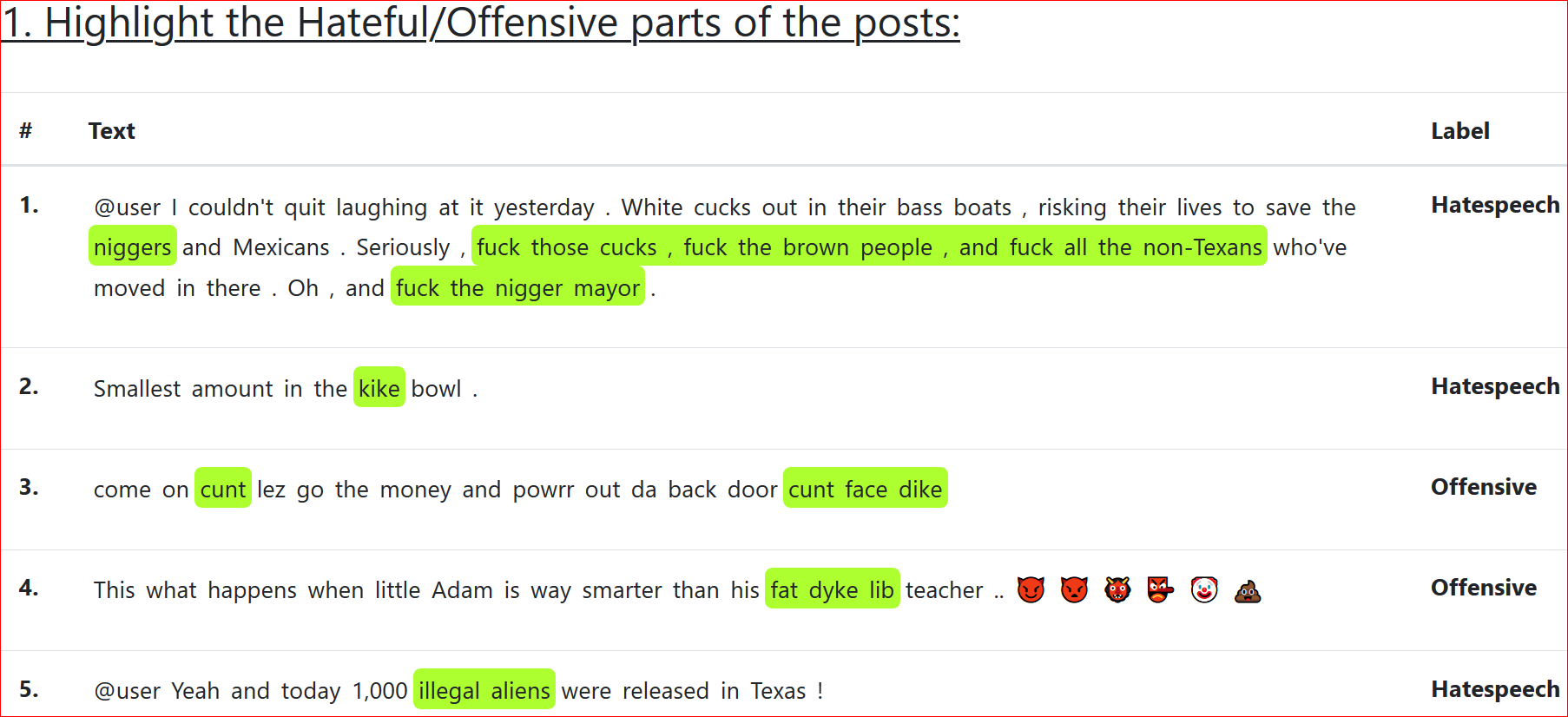}
 \caption{Rationale highlight. The annotators are asked to highlight the portions of the text that would justify the label.} 
 \label{fig:anno3}
\end{figure}

\begin{figure}[!tbh]
 \includegraphics[width=\linewidth]{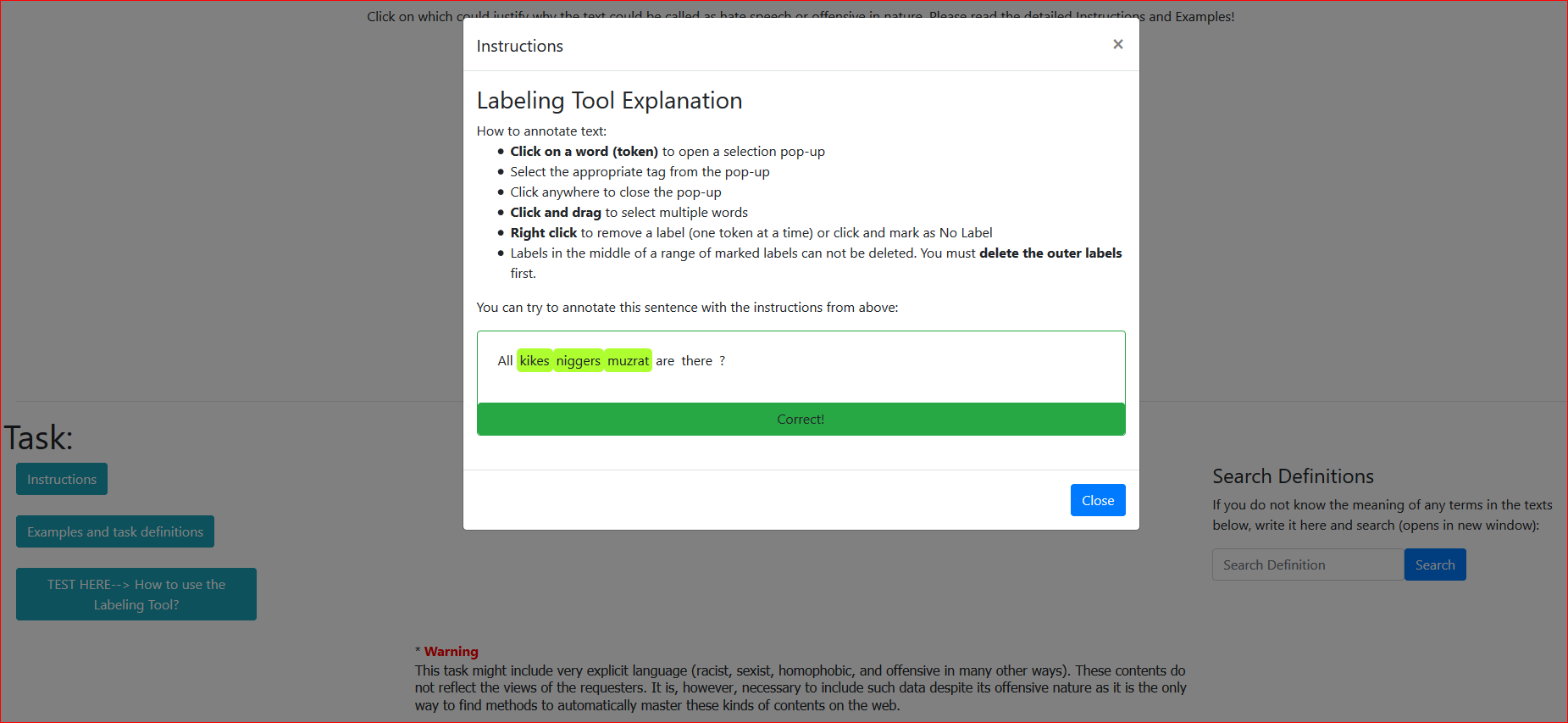}
 \caption{Highlight testing. The annotators are provided further instructions on how to highlight the texts.} 
 \label{fig:anno4}
\end{figure}

We divided the interface into two parts. First, as shown in Figure~\ref{fig:anno1}, we classified the text as hate speech, offensive, or normal, along with the targets in the text. Second, in Figure~\ref{fig:anno3}, we asked the annotators to highlight the portions of the text that could justify the label given to the text (hate speech/offensive). In order to help the annotators with the task, we provided them with multiple example annotations and highlights. We also provided them with sample test cases (as shown in Figure~\ref{fig:anno4}) to test out the highlight system.

\begin{table*}[!tbh]
\centering
\caption{This table represents the different hyper-parameter variations that we tried while tuning this model.}
\label{tab:hyperparameter}
\begin{tabular}{|l|l|l|l|l|}
\hline
Hyper-parameters &
  BERT &
  BiRNN &
  \begin{tabular}[c]{@{}l@{}}BiRNN-\\ Attention\end{tabular} &
  CNN-GRU \\ \hline
\begin{tabular}[c]{@{}l@{}}No. of hidden \\ units in sequential layer\end{tabular} &
  -NA- &
  64, 128 &
  64, 128 &
  -NA- \\ \hline
Sequential layers type        & -NA-           & LSTM,GRU       & LSTM,GRU       & GRU            \\ \hline
Train embedding layer         & -NA-           & True, False    & True, False    & True, False    \\ \hline
Dropout after embedding layer & -NA-           & 0.1.0.2,0.5    & 0.1,0.2,0.5    & 0.1,0.2,0.5    \\ \hline
\begin{tabular}[c]{@{}l@{}}Dropout after fully connected \\ layer\end{tabular} &
  0.1,0.2,0.5 &
  0.1,0.2,0.5 &
  0.1,0.2,0.5 &
  0.1,0.2,0.5 \\ \hline
Learning rate                 & 2e-4,2e-5,2e-6 & 0.1,0.01,0.001 & 0.1,0.01,0.001 & 0.1,0.01,0.001 \\ \hline
\multicolumn{5}{|l|}{\textbf{For supervised part}}                                                \\ \hline
Attention lambda ($\lambda$) &
  \begin{tabular}[c]{@{}l@{}}0.001,0.01,0.1,\\ 1,10,100\end{tabular} &
  -NA- &
  \begin{tabular}[c]{@{}l@{}}0.001,0.01,0.1,\\ 1,10,100\end{tabular} &
  -NA- \\ \hline
Number of supervised heads ($x$)    & 1,6,12         & -NA-           & -NA-           & -NA-           \\ \hline
\end{tabular}
\end{table*}

\section{Attention supervision in BERT}

In each encoder layer of BERT, an attention head computes key and query vectors to generate the \textit{attention values} for each token, based on other tokens in the sequence.  These attention values multiplied with the input token representations generates the weighted encoded representation of the token. This way we get a representation of each token from each of the 12 heads. The outputs of each head in the same layer are combined and run through a fully connected layer. Each layer is wrapped with a skip-connection and a layer normalization is applied after it. 

\begin{figure}[!tbh]
 \includegraphics[width=\linewidth]{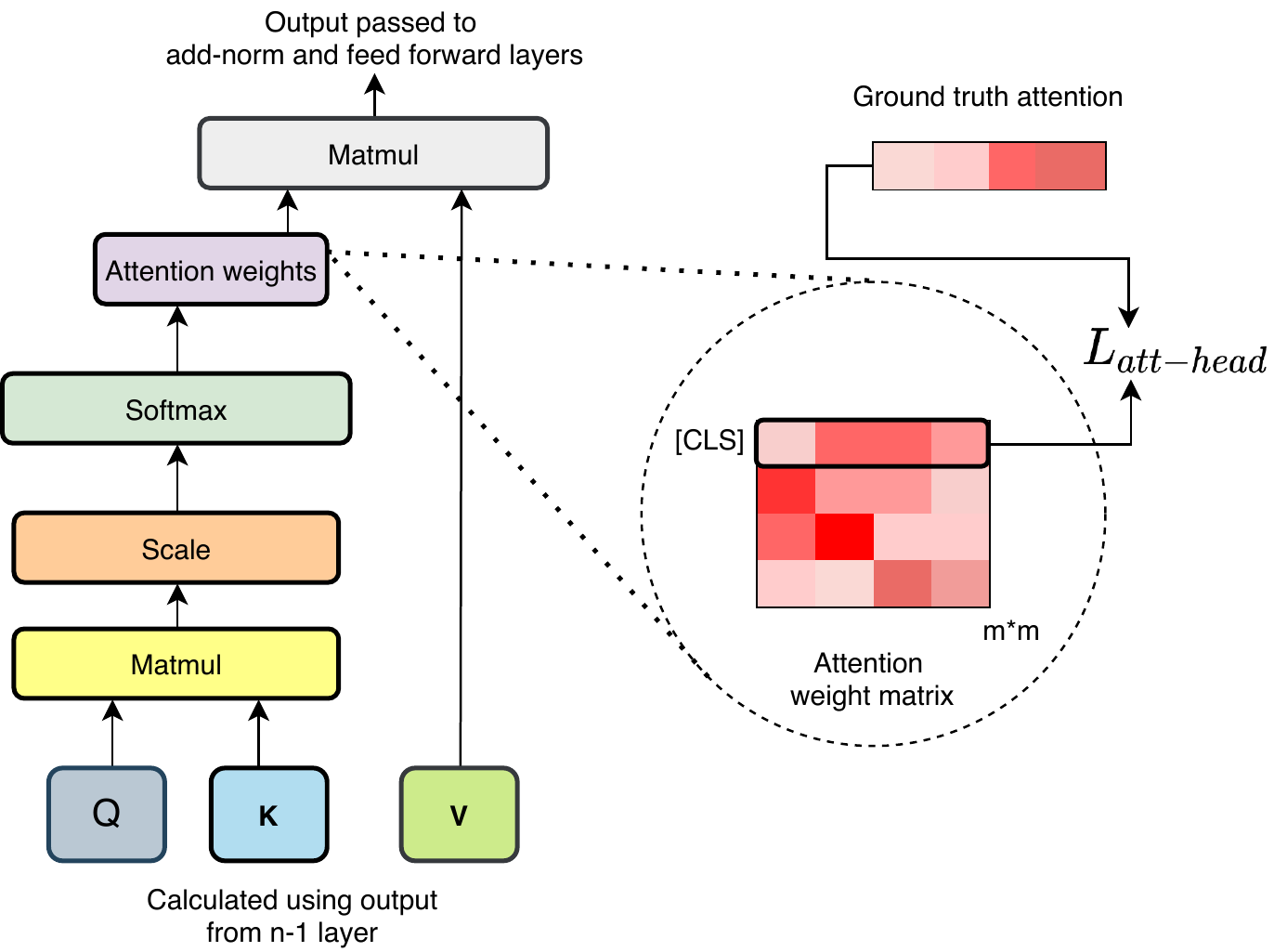}
 \caption{In this figure, we show attention supervision for a particular head in the $n$\textsuperscript{th} layer. For this work, we use the last layer for attention supervision. The number of heads for training is a hyper-parameter in our experiment.} 
 \label{fig:bert_supervised}
\end{figure}

For attention supervision, we use $x$ heads out of 12 heads in the last layer of BERT. We call these heads --- \textit{supervised heads}. For each supervised head, we use the attention weights corresponding to [CLS]\footnote{the first row in the $m*m$ attention weight matrix where $m$ is the number of tokens in the tokenized sentence.} and calculate the cross entropy loss with ground truth attention vector as shown in Figure \ref{fig:bert_supervised}. This ensures that the final weighted output corresponding to \textit{CLS} will give attention to words similar to the ground truth attention vector. Similarly, we do these steps for all \textit{supervised heads}. The final loss from the attention supervision is the average of the cross entropy loss from each supervised heads, which is further multiplied with the regulariser-$\lambda$. Other details about the finetuning is noted in \textit{BERT} section of the main paper.

\section{Hyper-parameters used}

Table~\ref{tab:hyperparameter} reports the hyper-parameters we tested for our systems.

\section{More examples}
Table~\ref{tab:more_examples} lists more examples corresponding to model predictions and rationales. In this example, BERT-HateXplain which uses token level rationales can attend better as compared to BERT. The prediction outcome is also correct for the BERT-HateXplain model. Wrong/incomplete attention (as shown) is one of the reason for incorrect predictions in BERT. For many of the false positives in BERT-HateXplain, the attended words are correct. In future, we plan to devise better mechanisms that can utilise attention for prediction.

\begin{table}[t]
\centering
\begin{tabular}{lll}
\toprule
Model & Text & Label \\
\midrule
Human Annotator & \hlg{I} \hlg{hate} \hlg{arabs} & HS \\ \midrule
BERT & I \hlg{hate} arabs & Normal \\
BERT-HateXplain & I \hlg{hate} \hlg{arabs} & HS \\ \midrule
\bottomrule
\end{tabular}
\caption{Example of the rationales predicted by different models compared to human annotators. The \hlg{green highlight} marks tokens that the human annotator and the model finds important for the prediction.}
\label{tab:more_examples}
\end{table}

\section{Annotations}
We looked into the quality of the fine-grained annotations as well. For this we computed the average pairwise Jaccard overlap between ground truth rationale annotations and compared them with the average pairwise overlap obtained through random annotation of the rationales. To generate the random rationale annotation, we chose 5 random tokens from a sentence as the rationale. For each sentence, we repeat this trial three times in order to denote 3 annotators. The average pairwise Jaccard overlap between the ground-truth annotations is 0.54, as compared to the random baseline of 0.36. This means that the annotators had more agreement on the token span annotations.

\end{document}